\documentclass[sigconf,nonacm]{acmart}

\usepackage{booktabs}
\usepackage{graphicx}
\usepackage{xurl}
\graphicspath{{../figures/}{./figures/}{./}}
\begin{document}

\title{OrchestraBench: Evaluating Multi-Agent Orchestration Failure
Modes, Recovery, and Decomposition Quality}

\author{Yidian Chen}
\affiliation{%
  \institution{Anote}
  \country{}
}
\author{Yingzi Gu}
\affiliation{%
  \institution{Anote}
  \country{}
}
\author{Natan Vidra}
\affiliation{%
  \institution{Anote}
  \country{}
}
\author{Spurthi Setty}
\affiliation{%
  \institution{Anote}
  \country{}
}
\author{Sharon Zheng}
\affiliation{%
  \institution{Anote}
  \country{}
}

\begin{abstract}
Multi-agent orchestration frameworks (AutoGen, LangGraph, CrewAI, Anthropic
Agents SDK) are moving from demos to production, yet they can only be compared
on \emph{task accuracy} --- not on \emph{reliability}. Existing agent benchmarks
measure whether a pipeline succeeds, but cannot diagnose \emph{why} it failed,
\emph{where} a cascade began, or \emph{which} routing decision caused the
breakdown. We present \textbf{OrchestraBench}, a benchmark that evaluates
how multi-agent systems \textbf{fail, recover, and decompose} as first-class,
comparable metrics. OrchestraBench contributes (i) a controlled,
seed-reproducible \textbf{failure-injection harness} over templated enterprise
workflows, (ii) \textbf{cascade-radius} and per-failure-mode recovery as primary
metrics, and (iii) a \textbf{routing-policy comparison} with bootstrap
confidence intervals and paired tests where applicable. Our first experiment
isolates the routing-mechanism question: on a
26-case gold-labelled diagnostic, a keyword/flag heuristic --- representative of
production rule-based routers --- scores \textbf{0\% on adversarial cases}
(misleading or missing surface flags), while a model-driven router that reasons
over intent scores \textbf{100\%}, matching the oracle on this diagnostic.
This indicates that the heuristic's blind spot is a property of the routing
\emph{mechanism}, not the diagnostic's inherent difficulty. Two further
experiments, run with a real Claude agent over a \textbf{verifiable arithmetic
dependency chain} as a controlled mechanism probe, isolate a reliability axis
existing benchmarks leave undermeasured. Across five MAST failure modes, failure
handling splits into \textbf{three tiers} --- a tool fault is fully recovered
(\textbf{1.0}), ambiguous delegation is \emph{partially} recovered
(\textbf{0.30}), and three latent/semantic modes never recover (\textbf{0.0}).
Two behaviors make this a property of the model rather than of the construct:
the tier \textbf{ordering survives} both reframing the identical computation as a
loan-approval workflow \emph{and} a three-model sweep (Sonnet / Opus / Haiku),
while absolute rates shift with context (tool $1.0\rightarrow0.7$, ambiguous
$0.30\rightarrow0.40$), and \textbf{retry does not
repair the latent modes} --- it reproduces the fault and only lengthens
time-to-detection, so detection/attribution, not blind retry, is the necessary
containment mechanism. As a corroborating structural signature, cascade radius
grows with pipeline depth (mean 0.9 to 4.7 across depths 3--7). A
policy-conditioned probe asks whether trusted-state repair can contain these
cascades; an ablation shows its apparent gain is largely the trusted-state signal
rather than autonomous detection, so we report it as a trusted-state probe, not a
deployable routing result. We frame these as controlled-chain mechanism probes,
not domain-workload claims (\S\ref{sec:limitations}).
\end{abstract}



\maketitle

\section{Introduction}
Multi-agent systems increasingly orchestrate several specialized agents ---
routers, retrievers, tool-callers, planners --- behind a single task. When such
a pipeline produces a wrong or low-quality answer in production, the failure is
often \emph{silent}: a misrouted task still yields a plausible-looking response.
Current benchmarks (AgentBench~\cite{agentbench}, OdysseyBench~\cite{odysseybench},
SWE-Bench~\cite{swebench}) report final task success and so cannot tell teams
\textbf{which} orchestration decision failed, \textbf{how far} an error
propagated, or \textbf{whether} intelligent routing was worth its added
complexity. This
blocks the most basic production decision: choosing and hardening an
orchestration framework on \emph{reliability} rather than ergonomics.

\paragraph{Contributions.}
\begin{enumerate}
  \item \textbf{OrchestraBench}, a benchmark that treats orchestration
  \emph{failure handling} as the unit of evaluation: controlled failure
  injection, per-failure-mode recovery, and cascade propagation.
  \item A \textbf{seed-reproducible} workflow + failure-injection harness
  (scenarios regenerable from seeds; aggregates recomputable from committed
  artifacts) with templated enterprise workflow families.
  \item A \textbf{routing-policy comparison} (Fixed, Heuristic, LLM, and Oracle)
  with bootstrap confidence intervals and paired significance testing where
  applicable, isolating the routing \emph{mechanism} effect on adversarial cases
  (Exp 1); the graded ``when does routing pay off'' comparison on larger
  workflow suites is future work.
\end{enumerate}

\paragraph{What we find.} On a controlled arithmetic chain, orchestration
failure handling splits into three tiers by fault type (tool faults recovered,
ambiguous delegation partially, three latent/semantic modes never). The result we
read as central is not the split itself but its \emph{robustness}: the tier
ordering survives reframing the identical computation as a business
loan-approval workflow while absolute rates move with context, and \emph{retry
does not repair the latent modes} --- evidence that failure handling is model
behavior under context, not an artifact of the construct. Cascade radius grows
with pipeline depth (0.9 to 4.7 across depths 3--7) as a supporting structural
signature (\S\ref{sec:limitations}).

\paragraph{Research questions.}
\begin{itemize}
  \item \textbf{RQ1 (routing value)}: When does intelligent routing (heuristic or
  LLM) actually beat a zero-decision baseline? \emph{(Partially: Exp 1 isolates
  the mechanism gap; a graded answer needs the larger suite.)}
  \item \textbf{RQ2 (mechanism)}: Is routing reliability a property of task
  \emph{difficulty}, or of the routing \emph{mechanism} (keyword/flag matching
  vs.\ reasoning over intent)? \emph{(Answered by Exp 1.)}
  \item \textbf{RQ3 (attribution)}: Can a pipeline failure be attributed to a
  specific failure mode \emph{and} stage, reproducibly across runs?
  \item \textbf{RQ4 (cascade)}: How far does a single seeded error propagate
  downstream, and which routing/recovery strategies contain it?
\end{itemize}

\section{Related Work}
\label{sec:related}
OrchestraBench sits among recent work that \emph{observes}, \emph{attributes}, or
\emph{proposes} fixes for multi-agent failure; we differ by making controlled
injection, cascade radius, and routing-policy mechanism the unit of analysis
(Table~\ref{tab:related}).

\begin{table*}[t]
\centering
\caption{OrchestraBench versus closely related work.}
\label{tab:related}
\small
\begin{tabular}{@{}p{3.4cm}p{6.0cm}p{6.0cm}@{}}
\toprule
\textbf{Prior work} & \textbf{What it does} & \textbf{How OrchestraBench differs} \\
\midrule
MAST~\cite{mast} (14 failure modes, 1{,}600+ traces, $\kappa{=}0.88$) &
Empirically-grounded \emph{taxonomy} of observed MAS failures &
We \emph{inject} that taxonomy under seed-controlled conditions and measure
recovery + cascade \emph{per mode} --- MAST observes, we intervene \\
MAS-FIRE~\cite{masfire} &
Fault injection + reliability evaluation; intra-/inter-agent fault taxonomy &
\textbf{Closest prior work} (\S\ref{sec:novelty}). We differ on emphasis: cascade
\emph{radius across pipeline depths} + \emph{routing-policy comparison} as the
unit of analysis, not reliability scoring alone \\
TraceElephant~\cite{traceelephant} (220 annotated traces) &
Benchmark for post-hoc failure attribution in MAS &
Attribution is observational; we add controlled seeding + cascade radius as
\emph{primary} metrics \\
Agents Failure Attribution~\cite{attribution} (Who\&When; 127 MAS) &
Post-hoc attribution to the responsible agent/step &
Same: attribution is a means here, not the product \\
Orchestration-pattern benchmark~\cite{orchpattern} (10k SEC filings) &
Compares sequential / parallel / hierarchical / reflexive architectures on a
cost-accuracy Pareto &
Architecture comparison on accuracy/cost; we evaluate \emph{failure handling} and
routing \emph{mechanism} --- complementary \\
AdaptOrch~\cite{adaptorch} &
Task-adaptive orchestration \emph{framework} (+12--23\%) &
A method; OrchestraBench is the benchmark that would evaluate it \\
From Spark to Fire~\cite{sparkfire} &
Errors cascade exponentially in pipelines &
We operationalize this into a measurable cascade-radius benchmark \\
\bottomrule
\end{tabular}
\end{table*}

\paragraph{Motivating data point.} \emph{Beyond the
Strongest}~\cite{beyondstrongest} reports that on GPQA-Diamond at least one agent
was correct in \textbf{95.5\%} of cases, yet orchestration reached only
\textbf{87.4\%} --- i.e.\ orchestration \emph{discards} $\sim$8 points of
individually-recoverable correctness. OrchestraBench is built to attribute
exactly this loss.

\paragraph{The gap.} Existing work either \emph{observes} failures
(MAST~\cite{mast}), \emph{attributes} them post-hoc
(TraceElephant~\cite{traceelephant}, \cite{attribution}), or \emph{proposes}
better orchestration (AdaptOrch~\cite{adaptorch}). The one work that also
\emph{injects} (MAS-FIRE~\cite{masfire}) scores reliability but does not make
cascade radius or routing-policy mechanism its unit of analysis. OrchestraBench's
contribution is a \textbf{controlled, reproducible injection harness that
measures recovery and cascade containment as first-class metrics, compared across
routing policies.}

\subsection{Positioning Relative to Closest Prior Work}
\label{sec:novelty}
An honest audit (the space moved fast in early 2026) surfaces one direct overlap
and our response:
\begin{itemize}
  \item \textbf{MAS-FIRE~\cite{masfire} already performs controlled fault
  injection + reliability evaluation}, overlapping our Exp 2/3 substantially. Our
  differentiation is \emph{not} ``we inject failures'' (they do too); it is (a)
  the \textbf{behavioral characterization} of failure handling --- the sharp
  tool-vs-latent recovery split, its persistence under domain reframing, and
  retry's inability to repair latent faults; (b) \textbf{cascade radius as a
  stages-traversed metric across pipeline depths}, which MAS-FIRE does not
  quantify (we report it honestly as partly structural on our verifiable chain);
  and (c) the \textbf{routing-mechanism result} of Exp 1. In our reading,
  MAS-FIRE does not quantify cascade depth as a stages-traversed metric and does
  not vary routing policy; its axes are topology (MetaGPT / Table-Critic /
  CAMEL), fault-type (15), and model. Our per-routing-policy containment is a
  probe rather than a headline (the \S\ref{sec:policyprobe} ablation shows the LLM policy's gain is
  largely the trusted-state signal). OrchestraBench therefore complements
  MAS-FIRE by adding a behavioral, depth-resolved view of the same failure
  taxonomy.
  \item \textbf{Attribution benchmarks~\cite{traceelephant,attribution} are
  observational or post-hoc} --- a different instrument from seeded,
  reproducible injection and recovery.
  \item \textbf{Internal overlap: IntentBench~\cite{intentbench}} evaluates
  whether the agent's \emph{understanding of intent} is correct (intent-spec vs.\
  syntactic tool-call correctness); OrchestraBench evaluates which
  \emph{orchestration action} to take given a correct understanding and how those
  decisions cascade. We measure the \textbf{decision policy and its failure
  propagation}, not intent correctness --- complementary ends of the pipeline,
  no measurement overlap.
\end{itemize}

\section{The OrchestraBench Benchmark}
OrchestraBench is released as a reusable artifact --- a seed-controlled workflow
generator, the \texttt{FailureInjector}, the routing-policy suite, and the
analysis scripts --- paired with the initial mechanism-probe results reported
here; the broad multi-domain evaluation it is designed to host is future work.

\textbf{Workflow suites.} Synthetically generated from templated task graphs
(stages, inter-task dependencies, per-task complexity score),
author-screened for realism. Three enterprise families: finance approval
(4 stages), HR onboarding (5 stages), DevOps deployment (5 stages).

\textbf{Routing decision space.} Each task is routed to one of four actions:
\texttt{DIRECT\_TOOL}, \texttt{CODE\_EXECUTION}, \texttt{DECOMPOSE},
\texttt{REASON\_ONLY}.

\textbf{Gold labels.} For the failure-recovery, cascade, and decomposition
experiments (Exp~2--4), ground truth is \emph{objective}: an exact-match check
against the verifiable arithmetic result, requiring no subjective annotation.
The Exp~1 routing diagnostic (26 cases) is author-labelled from unambiguous task
intent; a two-annotator protocol with reported inter-annotator agreement
($\kappa$) remains future work for the larger workflow-suite evaluation
(\S\ref{sec:limitations}).

\textbf{Failure injection.} Failure scenarios are seeded \textbf{deterministically}
on top of the suites via the \texttt{FailureInjector}, so each scenario is
reproducible from a seed.

\textbf{Metrics.} Primary reported metrics are routing accuracy,
per-failure-mode recovery rate, \textbf{cascade radius}, time-to-detection,
recovery completeness, and decomposition delegation fidelity. The harness also
records routing macro-F1, per-class routing metrics, success rate, latency, cost,
and orchestration efficiency for larger suite-level comparisons.

\section{Experimental Setup}
\label{sec:setup}
\begin{table}[t]
\centering
\caption{Routing policies under comparison.}
\label{tab:policies}
\small
\begin{tabular}{@{}llp{3.4cm}@{}}
\toprule
\textbf{Policy} & \textbf{Type} & \textbf{What it tests} \\
\midrule
Fixed & Baseline & Zero-intelligence lower bound (always one route) \\
Heuristic & Baseline & Rule/keyword routing on complexity + flags \\
LLM-as-Router & Test & Model decides routing per task by reasoning over
description + metadata \\
Retry(heuristic) & Test & Adds automatic retry --- does retry alone contain
cascades? \\
Oracle & Ceiling & Routes to the gold label (upper bound) \\
\bottomrule
\end{tabular}
\end{table}
Reported confidence intervals use 95\% bootstrap CIs. For the paired
comparisons (Exp 4; the policy ablation) we report an \emph{exact} two-sided
sign-flip permutation $p$ ($\alpha = 0.05$): with small, often-separated paired
samples the percentile-bootstrap $p$ is anti-conservative (it collapses to $0$
when every paired difference shares a sign), so the permutation test is the
honest instrument. Per-mode aggregates are $n{=}30$ (Exp 2) and $n{=}120$/depth
(Exp 3), pooling the baseline policies. The fully-deterministic cells (e.g.\
context pollution at final-task success 0.0) remain point masses regardless of
$n$ --- a corrupted latent state always fails the final task --- so their CIs are
zero-width by construction; the stochastic cells (ambiguous delegation; the
domain framings) carry informative CIs (\S\ref{sec:limitations}).

\section{Results}
\label{sec:results}
\subsection{Experiment 1 --- Routing Policy Comparison (measured)}
\label{sec:exp1}
We isolate the routing-mechanism question on a focused diagnostic of
\textbf{26 expert-labelled gold cases} covering all four routing decisions: 16
\emph{aligned} (surface flags match intent) and 10 \emph{adversarial} (flags
missing or misleading, but the description's intent is unambiguous). The
model-driven router is Claude Sonnet 4.6 via forced structured (tool-use) output;
results are stable across 3 independent passes (78/78), and the prompt never sees
the gold label.

\begin{table}[t]
\centering
\caption{Experiment 1: routing accuracy by case type.}
\label{tab:exp1}
\small
\begin{tabular}{@{}lccc@{}}
\toprule
\textbf{Policy} & \textbf{Overall} & \textbf{Aligned} & \textbf{Adversarial} \\
\midrule
Fixed (no signal) & 23\% & 25\% & 20\% \\
Heuristic (keyword/flags) & 62\% & \textbf{100\%} & \textbf{0\%} \\
TF-IDF (reads description) & 92\% & 88\% & \textbf{100\%} \\
LLM-as-Router (Sonnet 4.6) & \textbf{100\%} & 100\% & \textbf{100\%} \\
Oracle (ceiling) & 100\% & 100\% & 100\% \\
\bottomrule
\end{tabular}
\end{table}

\begin{figure}[t]
\centering
\includegraphics[width=\linewidth]{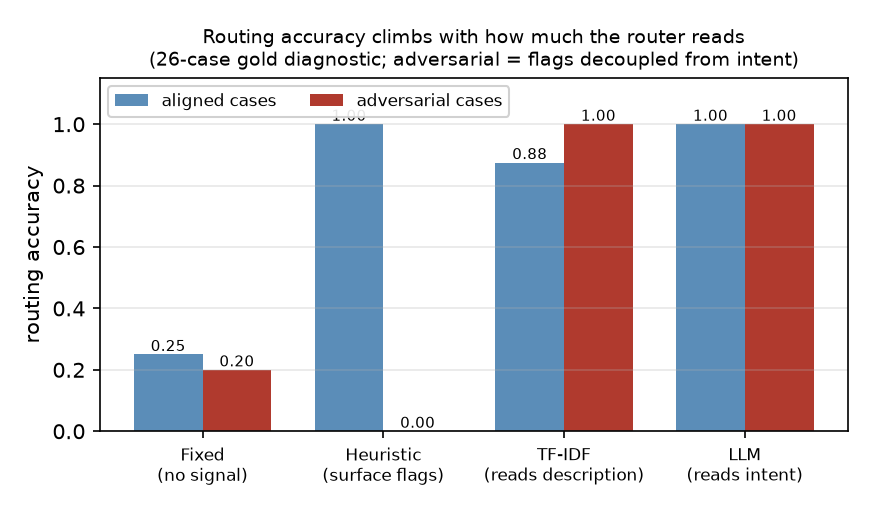}
\caption{Routing accuracy on the 26-case diagnostic climbs with how much of the
task the router reads. The flag heuristic's aligned/adversarial cliff (100\%/0\%)
is flattened by a description-reading TF-IDF router (88\%/100\%) with no model,
isolating the heuristic's blind spot as a mechanism property.}
\Description{A bar chart comparing routing accuracy for Fixed, Heuristic, TF-IDF, and LLM routers on aligned versus adversarial cases. Fixed is low on both, Heuristic is perfect on aligned but zero on adversarial, and both TF-IDF and LLM are near-ceiling on adversarial cases.}
\label{fig:exp1ladder}
\end{figure}

\textbf{Finding.} The heuristic is perfect on aligned cases but collapses to
\textbf{0\%} on adversarial ones; the model-driven router recovers \textbf{all} of
them (0\% $\to$ 100\%; Wilson 95\% CI on the 10 adversarial cases $[0.00, 0.28]$
vs.\ $[0.72, 1.00]$), matching the oracle on this diagnostic. The contrast is not
merely keyword-versus-LLM: a \textbf{mid-strength TF-IDF router} that simply reads
the task \emph{description} against fixed per-route prototypes --- no model, no
learned weights --- already recovers every adversarial case (100\%), while giving
back a few aligned cases where the surface flags were in fact the cleaner signal
(88\%). The takeaway is not a smooth ``more reading is better'' gradient:
surface-flag matching actually scores \emph{below} the no-signal baseline on
adversarial cases (0\% vs.\ 20\%), since misleading flags are worse than none. The
sharp line is \emph{what} the router reads --- any router that reads the task
\emph{description} clears the adversarial set (TF-IDF and LLM both 100\%), while
flag-matching alone collapses to 0\% (Fig.~\ref{fig:exp1ladder}). Because a shallow text router already clears
the adversarial set, the heuristic's collapse is a property of the routing
\textbf{mechanism} --- matching surface flags that, by construction, decouple from
intent --- not of inherent task difficulty. The adversarial set is deliberately a
worst case for surface-flag matching; we read Exp 1 as isolating \emph{when}
keyword routing fails (flags decoupled from intent), not as a general
difficulty-graded benchmark. Larger workflow-suite evaluation is the intended
setting for a graded comparison once the diagnostic no longer saturates.

\subsection{Experiment 2 --- Failure Injection and Recovery (real Claude
measured run; core)}
\textbf{Design.} Five MAST failure modes --- \emph{ambiguous delegation},
\emph{tool-invocation error}, \emph{context pollution}, \emph{conflicting
sub-agent outputs}, \emph{premature action} --- injected at the \textbf{prompt
level} into a verifiable arithmetic dependency chain executed by a \textbf{real
Claude agent} (Sonnet 4.6), under fixed, heuristic, and retry policies. Real
recovery and cascade are measured by exact-match against ground truth; the
measured-run module is included in the repository.

\begin{table}[t]
\centering
\caption{Experiment 2: per-mode final-task success and cascade radius, arithmetic
chain vs.\ loan-approval domain (real Claude, $N{=}150$ each, $n{=}30$/mode;
95\% bootstrap CI).}
\label{tab:exp2}
\small
\begin{tabular}{@{}lcccc@{}}
\toprule
& \multicolumn{2}{c}{\textbf{Arithmetic}} & \multicolumn{2}{c}{\textbf{Domain}} \\
\cmidrule(lr){2-3}\cmidrule(lr){4-5}
\textbf{Failure mode} & \textbf{Success} & \textbf{Cascade} & \textbf{Success} &
\textbf{Cascade} \\
\midrule
tool\_invocation\_error & 1.00 & 0.00 & 0.70 & 0.60 \\
ambiguous\_delegation & 0.30 & 1.40 & 0.40 & 1.20 \\
context\_pollution & 0.00 & 2.00 & 0.00 & 2.00 \\
conflicting\_outputs & 0.00 & 2.00 & 0.00 & 2.00 \\
premature\_action & 0.00 & 2.00 & 0.00 & 2.00 \\
\bottomrule
\end{tabular}
\end{table}

\textbf{Real measured findings (Sonnet 4.6, $N{=}150$, $n{=}30$/mode).}
Failure handling splits into \textbf{three tiers}. (i) The tool-invocation mode
is \textbf{fully recovered} (recovery 1.0, cascade radius 0) --- the agent
recomputes by hand when the tool fails. (ii) \textbf{Ambiguous delegation is
partially recovered} (final-task success \textbf{0.30 [0.13, 0.47]}, cascade
radius 1.40) --- with $n{=}30$ the agent infers the intended operation about a
third of the time, a genuinely \emph{stochastic} cell rather than a point mass.
(iii) The remaining three latent/semantic modes (context pollution, conflicting
outputs, premature action) \textbf{never recover} (final-task success
\textbf{0.0}) and cascade to every downstream stage. Critically, the retry policy
\textbf{does not recover the latent modes}: retrying while the failure is still
present reproduces the error and only \textbf{lengthens time-to-detection} ---
retry repairs retryable tool faults but not failures needing attribution, state
repair, or semantic validation (RQ3/RQ4). The stochastic cells now carry
informative CIs (ambiguous 0.30 [0.13, 0.47]); the fully-deterministic 0.0 and 1.0
cells remain point masses by construction (\S\ref{sec:limitations}).

\begin{figure}[t]
\centering
\includegraphics[width=\columnwidth]{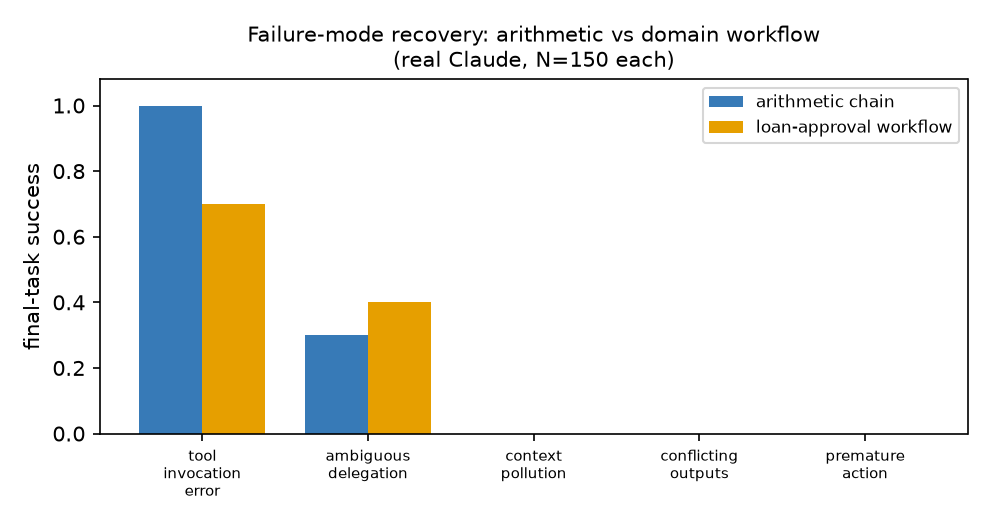}
\caption{Experiment 2 --- failure-mode final-task success, arithmetic chain vs.\
loan-approval workflow; the failure-mode \emph{ordering} is robust across framings
while absolute rates shift (real Claude, $N{=}150$ each).}
\Description{A grouped bar chart comparing arithmetic-chain and loan-approval variants across five failure modes. Tool faults remain the most recoverable, ambiguous delegation shifts modestly, and the three latent semantic modes remain at zero in both framings.}
\label{fig:exp2}
\end{figure}

\textbf{Domain-grounded validation (loan-approval workflow, $N{=}150$).} To test
whether these signatures are an artifact of the abstract arithmetic chain, we
re-ran the \emph{identical} failure injection on a domain-grounded variant: the
same verifiable computation reframed as a four-role loan-approval pipeline
spanning Intake Officer, Risk Analyst, Compliance Officer, and Approval
Manager, with each stage prompted in business terms. \textbf{The core ordering
holds}: context pollution, conflicting outputs, and premature action still fail
completely, while tool faults remain the most recoverable
(Figure~\ref{fig:exp2}). Crucially, the \textbf{absolute rates shift with
framing}: ambiguous delegation rises from \textbf{0.30} to \textbf{0.40} (the
business-role context helps the agent infer the intended operation) while
tool-invocation recovery falls from \textbf{1.00} to \textbf{0.70}. This
framing-sensitivity is evidence of model behavior under context, not a purely
structural tautology, and directly addresses the construct-validity concern
(\S\ref{sec:limitations}).

\textbf{Cross-model check (Sonnet 4.6, Opus 4.8, and Haiku 4.5; $N{=}150$ each).}
To test whether the tier structure is specific to one model, we re-ran the
arithmetic-chain probe on three Claude tiers spanning the weak/mid/strong
capability range (Table~\ref{tab:crossmodel}). The core structure is
\textbf{model-invariant}: tool faults are fully recovered (1.0) and the two
catastrophic latent modes (context pollution, premature action) never recover
(0.0) in \emph{all three} models. As with the domain reframing, only the
stochastic ambiguous-delegation rate moves with the model (0.10--0.33) without
changing the ordering (the Sonnet re-run, 0.33, matches the main-table 0.30
within $n{=}30$ noise). The tool-vs-latent gap --- the paper's central finding
--- thus holds across capability tiers, not just Sonnet.

\begin{table}[t]
\centering
\caption{Cross-model check on the arithmetic chain across three Claude tiers
(Sonnet 4.6, Opus 4.8, Haiku 4.5; $N{=}150$ each, $n{=}30$/mode). The
tool-vs-latent structure is invariant; only the stochastic ambiguous-delegation
rate moves with the model.}
\label{tab:crossmodel}
\small
\begin{tabular}{@{}lccc@{}}
\toprule
\textbf{Failure mode} & \textbf{Sonnet 4.6} & \textbf{Opus 4.8} & \textbf{Haiku 4.5} \\
\midrule
tool\_invocation\_error & 1.00 & 1.00 & 1.00 \\
ambiguous\_delegation & 0.33 & 0.10 & 0.20 \\
context\_pollution & 0.00 & 0.00 & 0.00 \\
conflicting\_outputs & 0.00 & 0.10 & 0.00 \\
premature\_action & 0.00 & 0.00 & 0.00 \\
\bottomrule
\end{tabular}
\end{table}

\subsection{Experiment 3 --- Cascade Propagation Depth (real Claude measured run,
$N{=}750$; core)}
\textbf{Design.} Inject a single seeded error at stage 1 of variable-depth
pipelines (depths 3 through 7) and measure how far it propagates. \textbf{Cascade
radius} (downstream stages corrupted) is a differentiator vs.\ MAS-FIRE,
which has no stages-traversed metric.

\begin{table}[t]
\centering
\caption{Experiment 3: mean cascade radius by pipeline depth (real Claude,
$N{=}750$; latent-mode pool, $n{=}120$/depth; 95\% bootstrap CI). Latent cascade
radius grows monotonically with depth: the three catastrophic modes track depth
$-$ 2 exactly, while ambiguous delegation runs lower as it partially recovers.}
\label{tab:exp3}
\small
\begin{tabular}{@{}lcc@{}}
\toprule
\textbf{Depth} & \textbf{Latent cascade radius} & \textbf{Tool cascade radius} \\
\midrule
3 & 0.93 [0.88, 0.97] & 0.00 \\
4 & 1.85 [1.75, 1.93] & 0.00 \\
5 & 2.80 [2.65, 2.93] & 0.00 \\
6 & 3.63 [3.43, 3.83] & 0.00 \\
7 & 4.67 [4.42, 4.88] & 0.00 \\
\bottomrule
\end{tabular}
\end{table}

\begin{figure}[t]
\centering
\includegraphics[width=\columnwidth]{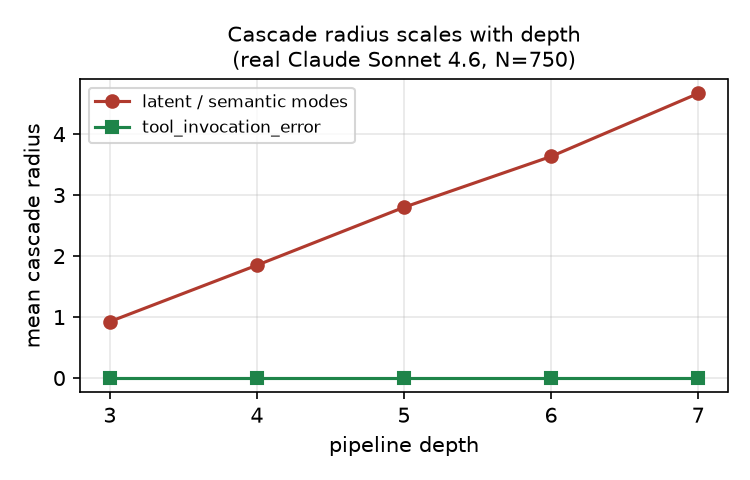}
\caption{Experiment 3 --- mean cascade radius vs.\ pipeline depth; latent/semantic
modes scale 0.93 $\to$ 4.67 across depths 3--7 while tool faults stay at
0 (real Claude, $N{=}750$).}
\Description{A two-line plot of mean cascade radius versus pipeline depth. The latent or semantic curve rises steadily from just under 1 at depth 3 to nearly 4.7 at depth 7, while the tool-invocation curve stays at zero throughout.}
\label{fig:exp3}
\end{figure}

\textbf{Real measured findings (Sonnet 4.6, $N{=}750$, $n{=}120$/depth).} Pooled
over the latent modes, \textbf{mean cascade radius grows monotonically with
depth: 0.93 $\to$ 1.85 $\to$ 2.80 $\to$ 3.63 $\to$ 4.67 across depths 3 / 4 / 5 /
6 / 7} (Table~\ref{tab:exp3}, Figure~\ref{fig:exp3}), an almost-linear
$\sim$0.9-per-stage trend now resolved at five depths with non-degenerate CIs.
The structure is sharp per mode: the three catastrophic modes (context pollution,
conflicting outputs, premature action) corrupt \emph{every} downstream stage
(cascade $=$ depth $-$ 2 exactly), while ambiguous delegation runs consistently
lower (0.7 / 1.4 / 2.2 / 2.5 / 3.7) because it partially recovers, and
tool-invocation errors stay at 0 at every depth (the agent recomputes by hand).
\textbf{This quantifies a stages-traversed signature MAS-FIRE leaves unmeasured};
on the verifiable chain the deterministic modes' growth is partly structural
(\S\ref{sec:limitations}), so we read the depth-scaling as corroborating rather
than as the paper's headline.

\subsection{Experiment 4 --- Decomposition Quality (real Claude measured run,
$N{=}40$; secondary)}
For composite tasks $(a \mathbin{op} b) \mathbin{op} (c \mathbin{op} d)$ with a
canonical 3-step gold decomposition, we score the produced decomposition against
gold --- \textbf{delegation fidelity} (gold sub-results recovered),
\textbf{granularity error}, \textbf{wasted sub-tasks} --- comparing a
\texttt{monolithic} (single-step) vs.\ a \texttt{decompose} (planner) policy.

\begin{table}[t]
\centering
\caption{Experiment 4: decomposition quality, decompose vs.\ monolithic (real
Claude, $N{=}40$; 95\% bootstrap CI; exact permutation $p$ over the 20 shared
tasks). Zero-width CIs are degenerate by construction (\S\ref{sec:setup}).}
\label{tab:exp4}
\small
\setlength{\tabcolsep}{4pt}
\begin{tabular}{@{}lccc@{}}
\toprule
\textbf{Policy} & \textbf{Deleg.\ fidelity} & \textbf{Granularity err.} &
\textbf{Final correct} \\
\midrule
decompose & 1.00 [1.00, 1.00] & 0.00 [0.00, 0.00] & 1.00 \\
monolithic & 0.37 [0.33, 0.42] & 2.00 [2.00, 2.00] & 1.00 \\
\midrule
paired $\Delta$ (fidelity) & \multicolumn{3}{l}{$+0.63$ [0.58, 0.67], $p{=}1.9{\times}10^{-6}$} \\
paired $\Delta$ (gran.) & \multicolumn{3}{l}{$-2.0$, $p{=}1.9{\times}10^{-6}$} \\
\bottomrule
\end{tabular}
\end{table}

\textbf{Real measured findings (Sonnet 4.6, $N{=}40$).} The \texttt{decompose}
policy produces a faithful three-step decomposition (delegation fidelity
\textbf{1.00}, granularity error \textbf{0}), while \texttt{monolithic} reaches
the correct final answer but exposes no delegable sub-structure (fidelity
\textbf{0.37 [0.33, 0.42]}, granularity error \textbf{2}). This fidelity gap is
large and statistically significant under an exact sign-flip permutation test
over the 20 shared tasks: \textbf{$+0.63$, 95\% CI [0.58, 0.67], $p{=}1.9{\times}10^{-6}$}
(Table~\ref{tab:exp4}). Notably, \texttt{final\_correct} is \textbf{1.0 for both}
--- the arithmetic is easy enough that both get the answer; the discriminator is
the \textbf{decomposition structure}, not final correctness. For multi-agent
orchestration this is exactly the point: a monolithic agent can be \emph{right}
yet produce nothing a planner can delegate, audit, or recover from.

\subsection{Policy-conditioned containment probe (real Claude, $N{=}75 + N{=}225$)}
\label{sec:policyprobe}
Experiments 2--3 pool over baseline policies; we add \textbf{LLM-as-router} and
\textbf{Oracle} policies to ask whether the routing policy itself can govern
containment. \textbf{The LLM-policy semantics are an explicit modelling
assumption}: at the injected stage the router is given the trusted upstream value
and licensed to detect and correct the anomaly. We therefore treat the LLM row as
a strong self-correction probe rather than a clean estimate of deployable routing
without trusted state.

\begin{table}[t]
\centering
\caption{Policy-conditioned containment on latent failures (real Claude, $N{=}75$; 95\% bootstrap CI).}
\label{tab:policy}
\small
\begin{tabular}{@{}lcc@{}}
\toprule
\textbf{Policy} & \textbf{Latent recovery} & \textbf{Cascade radius} \\
\midrule
fixed / heuristic / retry & 0.08 [0.00, 0.25] & 1.83 [1.50, 2.00] \\
\textbf{LLM-as-router} & \textbf{0.83 [0.58, 1.00]} & \textbf{0.33 [0.00, 0.83]} \\
Oracle (ceiling) & 1.00 [1.00, 1.00] & 0.00 [0.00, 0.00] \\
\bottomrule
\end{tabular}
\end{table}

\begin{figure}[t]
\centering
\includegraphics[width=\columnwidth]{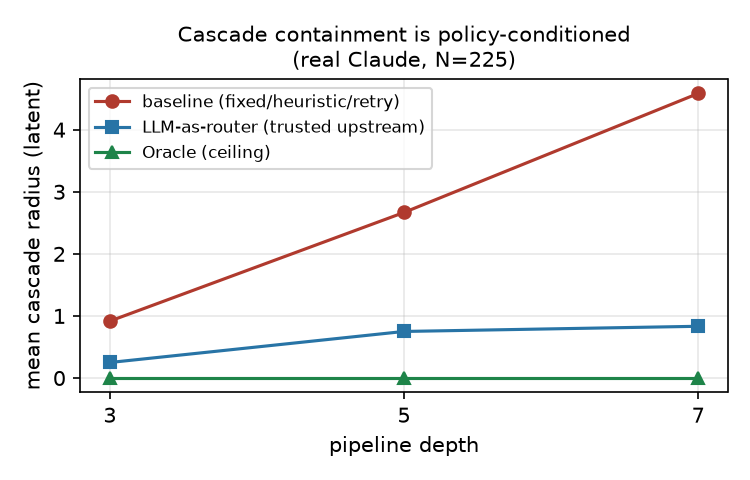}
\caption{Experiment 3 --- mean latent cascade radius vs.\ pipeline depth per
routing policy; baseline grows with depth while the trusted-state LLM probe stays
nearly flat and Oracle fully contains (real Claude, $N{=}225$).}
\Description{A three-line plot of latent cascade radius by pipeline depth for baseline, LLM-as-router, and Oracle policies. The baseline rises sharply with depth, the LLM line stays low and nearly flat, and Oracle remains at zero across all depths.}
\label{fig:policy}
\end{figure}

\textbf{Policy probe by depth ($N{=}225$).} Under this stronger trusted-state
semantics, the policy effect \emph{amplifies with pipeline depth}: baseline
cascade radius grows with depth (0.9 / 2.7 / 4.6 at depths 3 / 5 / 7), while the
LLM router stays nearly flat (0.25 / 0.75 / 0.83) and Oracle fully contains (0 /
0 / 0). At depth 7 the LLM router cuts cascade radius
\textbf{$\sim$5.5$\times$} versus the baselines. This suggests that containment
benefits can become larger in deeper pipelines, while the deployable-routing
version remains future work (Figure~\ref{fig:policy}). The policy CSVs are
committed under \texttt{data/measured/}.

\textbf{Ablation: the model, or the hint?} The rows above hand the router the
trusted upstream value --- a possible information leak. Re-running Exp 2 as an
independent, larger sweep ($N{=}180$; its LLM estimate 0.67 is a separate sample
from Table~\ref{tab:policy}'s $N{=}75$ value 0.83) with an added
\texttt{llm\_noupstream} policy (same self-correction, \emph{no} trusted-upstream
hint) collapses latent recovery from \textbf{0.67 [0.46, 0.83]} to
\textbf{0.08 [0.00, 0.21]}, down to the baseline level (paired drop
\textbf{0.58 [0.33, 0.79], $p{=}5.19{\times}10^{-4}$}, $n{=}24$ pairs). The LLM policy's containment is thus
\emph{mostly the trusted-upstream signal, not autonomous detection}: we read the
LLM column as a trusted-state probe, not evidence that an LLM router autonomously
contains latent cascades (\texttt{exp2\_ablation\_real.csv}).

\section{Discussion}
Experiment 1 establishes the paper's first reproducible claim: routing
reliability depends on \emph{mechanism}. Experiments 2 and 3 then deliver the
central behavioral finding: the agent's failure handling \emph{splits sharply by
fault type}. A tool-invocation fault is fully recovered (recovery 1.0) --- the
agent recomputes the value by hand --- ambiguous delegation is \emph{partially}
recovered (0.30, the agent sometimes infers the intended operation), and the
remaining three latent/semantic modes are \emph{never} recovered (0.0
final-task success); and \textbf{retry does not repair the latent modes}: it
reproduces the failure and only lengthens time-to-detection, so
\emph{detection and attribution}, not blind retry, is the necessary containment
mechanism. Crucially this split is \emph{model behavior, not a construct
artifact}: it survives reframing the identical computation as a loan-approval
workflow --- the failure-mode ordering persists while absolute rates shift with
context (Fig.~\ref{fig:exp2}). On top of this we quantify a structural signature
prior reliability benchmarks leave unmeasured: cascade radius grows with pipeline
depth (0.9 to 4.7 across depths 3--7, near-linear); on this verifiable chain that
growth is partly by construction (\S\ref{sec:limitations}), so we read it as
corroborating the depth-scaling of latent cascades rather than as the headline.
The policy-conditioned extension shows the potential value of trusted-state
repair, but its deployable interpretation is deliberately limited
(\S\ref{sec:limitations}). These are mechanism probes on a controlled chain;
domain-workflow validation is the next step.

\section{Limitations}
\label{sec:limitations}
\begin{itemize}
  \item The Exp 1 diagnostic is small (26 cases) and \emph{saturates} for a
  capable model --- it is a mechanism probe, not a difficulty benchmark; the
  larger workflow-suite results are needed for a graded comparison. Its labels
  are author-provided; a two-annotator protocol with reported $\kappa$ is future
  work for the larger suite.
  \item Workflows are synthetically generated; real-world pipeline validation is
  future work.
  \item \textbf{Model coverage.} The core Exp 2--4 numbers are on Claude Sonnet
  4.6. A three-model cross-check (Sonnet 4.6, Opus 4.8, and Haiku 4.5;
  Table~\ref{tab:crossmodel}) confirms the tool-vs-latent structure is
  \emph{model-invariant} across capability tiers --- tool faults fully recovered
  and the catastrophic latent modes never recovered in all three --- with only
  the stochastic ambiguous rate moving. A broader non-Claude sweep (e.g.\ GPT
  or Gemini) through the same cross-model harness remains future work.
  \item \textbf{Modest samples.} Per-mode aggregates are $n{=}30$ in
  Experiment~2 and $n{=}120$ per depth in Experiment~3; the fully-deterministic cells still yield zero-width
  CIs by construction (a corrupted latent state always fails the final task),
  while the stochastic cells now carry informative intervals. Larger, more
  diverse workflow suites remain future work.
  \item \textbf{Construct validity of Experiments~2 and~3.} The failure-recovery and cascade
  experiments run on a \emph{verifiable arithmetic dependency chain}, chosen for
  clean ground truth, not a domain workflow. In this construct cascade radius is
  \textbf{partly by design} --- a downstream stage consumes the upstream value,
  so a latent corruption propagates to $\sim$(depth $-$ 2) stages --- so we
  present Experiments~2 and~3 as \textbf{controlled mechanism probes}. The evidence that this
  measures model behavior rather than a tautology is the
  non-determinism we observe (ambiguous delegation stays below the structural
  maximum at every depth --- mean cascade 2.2 vs.\ 3 at depth 5 --- and recovers
  0.30 of the time) and the domain-grounded re-run
  (Figure~\ref{fig:exp2}): the failure-mode \emph{ordering} mostly persists
  across framings while absolute rates shift with context. Full finance, HR,
  and DevOps suite validation remains future work.
  \item \textbf{Single LLM, not a literal multi-agent system.} Experiments~2 and~3 execute one
  Claude agent over a staged chain with prompt-level fault injection ---
  simulating orchestration failure modes rather than wiring $N$ independent
  agents; a real multi-agent harness is future work.
  \item \textbf{LLM-policy semantics are a modelling assumption.} The
  policy-conditioned results define the LLM router as a stronger pass given the
  trusted upstream value and licensed to self-correct; the absolute LLM numbers
  depend on this choice. The Oracle (gold-repair ceiling) and baseline columns
  are unambiguous, but the LLM column should be read as a trusted-state
  self-correction probe, not yet as a deployable routing estimate. To isolate whether
  the LLM gain reflects the model detecting the fault versus being handed the trusted
  upstream, we ran an ablation policy (\texttt{llm\_noupstream}) that drops the
  trusted-upstream hint: latent recovery collapses to near-baseline (\S\ref{sec:policyprobe}), confirming
  the LLM column is a trusted-state probe, not autonomous detection.
\end{itemize}

\section{Ethics and Broader Impact}
OrchestraBench evaluates the \emph{reliability} of automated orchestration.
Improving failure attribution and cascade containment is intended to make
production AI systems \textbf{safer, more auditable, and cheaper to operate} ---
surfacing silent failures before deployment rather than after.
\begin{itemize}
  \item \emph{Leaderboard overfitting}: a public reliability benchmark can
  incentivize tuning to the injected failure modes. Mitigation: seed-controlled,
  \textbf{held-out} failure scenarios, and reporting cost alongside accuracy
  where deployment-cost comparisons are made.
  \item \emph{Dual use}: cataloguing how orchestrators fail could inform
  adversarial prompting. Mitigation: the injected modes are already public (MAST);
  we add measurement, not new attack surface.
  \item \emph{Over-trust}: a high score must not be read as production safety. We
  state scope limits (synthetic workflows; mechanism probes) explicitly.
\end{itemize}
\textbf{AI-use disclosure.} Generative AI tools were used for limited editorial
and coding assistance; all results, citations, and claims were checked against
committed code, data, and cited records.

No human-subjects data is used; all workflows are synthetic and include no real
applicants, employees, customers, or proprietary records.

\section{Reproducibility}
Failure scenarios are regenerable from fixed seeds, and every reported aggregate
is recomputable from committed artifacts. The Exp 1 reproduction script
regenerates the \S\ref{sec:exp1} table from the committed gold set (offline baselines need
no key; the LLM row runs when the Anthropic API key is set), and
the offline baseline script recomputes the Fixed, Heuristic, and TF-IDF rows
and Fig.~\ref{fig:exp1ladder} fully offline. A model-agnostic cross-model
harness re-runs the five-mode probe on any set of models into a pooled
comparison table. The measured
analysis script recomputes every Exp 2/3/4 and policy-probe number above
(per-mode recovery, cascade radius by depth, decomposition fidelity, and
trusted-state ablation) with bootstrap 95\% CIs and exact paired permutation
tests, directly from the committed CSVs (offline, no key).
Code is Apache-2.0; data and failure scenarios are seeded; the full test suite
(151 tests) runs offline on Python 3.10/3.11/3.12. Total real-LLM cost across the
project is $\approx\$4$ (Sonnet 4.6; short arithmetic prompts), so reproduction
is negligible.

\section{Conclusion}
OrchestraBench reframes multi-agent evaluation from ``did the task succeed?'' to
``did the orchestrator route, recover, and decompose reliably?'' Experiment 1
delivers a clean, reproducible result --- model-driven routing closes a
0\%$\to$100\% adversarial gap that keyword routing cannot --- while the
failure-injection experiments surface, on a controlled chain, the reliability
axis production teams most need: a sharp three-tier fault split (tool faults
fully recovered, ambiguous delegation partially recovered, three semantic faults
never recovered and beyond the reach of retry) that persists under domain
reframing, with cascade radius growing with pipeline depth (0.9 to 4.7 across
depths 3--7) as a corroborating structural signature.


\bibliographystyle{ACM-Reference-Format}
\bibliography{references}

\end{document}